\documentclass[runningheads]{llncs}
\usepackage{graphicx}
\usepackage{algorithm}
\usepackage[noend]{algcompatible}

\usepackage{adjustbox}
\begin{document}
\title{Contrastive learning for unsupervised medical image clustering and reconstruction}
\titlerunning{Contrastive autoencoder for deep clustering}
%
\author{Matteo Ferrante\inst{1} \and
Tommaso Boccato \inst{1} \and
Andrea Duggento\inst{1}\and
Simeon Spasov \inst{2} \and
Nicola Toschi\inst{3}}
\authorrunning{Ferrante et al.}
%
\institute{Department of Biomedicine and Prevention, University of Rome Tor Vergata, Rome, Italy
\and
Department of Computer Science and Technology University of Cambridge, UK 
\and
Department of Radiology, Athinoula A. Martinos Center for Biomedical Imaging, Boston, MA, USA \\}
\maketitle              
\begin{abstract}
The lack of large labeled medical imaging datasets, along with significant inter-individual variability compared to clinically established disease classes, poses significant challenges in exploiting medical imaging information in a precision medicine paradigm, where in principle dense patient-specific data can be employed to formulate individual predictions and/or stratify patients into finer-grained groups which may follow more homogeneous trajectories and therefore empower clinical trials. In order to efficiently explore the effective degrees of freedom underlying variability in medical images in an unsupervised manner, in this work we propose an unsupervised autoencoder framework which is augmented with a contrastive loss to encourage high separability in the latent space. The model is validated on (medical) benchmark datasets. As cluster labels are assigned to each example according to cluster assignments, we compare performance with a supervised transfer learning baseline. Our method achieves similar performance to the supervised architecture, indicating that separation in the latent space reproduces expert medical observer-assigned labels. The proposed method could be beneficial for patient stratification, exploring new subdivisions of larger classes or pathological continua or, due to its sampling abilities in a variation setting, data augmentation in medical image processing.
\end{abstract}

\section{Introduction}

In current medical practice, the difficulties in specifically targeting any disease process are rooted in recent evidence showing that current diagnostic categories do not actually represent a single disease, but rather heterogeneous clinical syndromes underpinned by different pathogenic mechanisms.
Today, we still do not know how heterogeneity in organ function and anatomy are linked to this clinical variability and to the risk of following different “trajectories”.
This represents a drawback in personalized diagnosis and therapy, which is commonly based on a “one size fits all" approach. This lack of understanding of the basic mechanisms underlying (multimorbid) syndromes is a major roadblock in fostering today’s P4\footnote{P4: predictive, preventative, personalized, participatory} medicine paradigm as well as in designing cost- and discovery- efficient clinical trials. For these reasons, there is significant interest in developing unsupervised methods which can discover clinical subtypes (or, more generally, more fine-grained patient strata ) in patient populations based on patient data \cite{Alzeihmer}. In this paper, we propose an unsupervised framework for image-based patient stratification based on an autoencoder network. We augment the reconstruction loss of the autoencoder with a contrastive learning component inspired by \cite{simclr,simclr2} to encourage better separation of the latent space \cite{contrastive,deep_review,dino,byol}. In the first stage of training (warmup), we focus on learning structured latent representations, while in the second stage we fine-tune the decoder for reconstruction. Using a simple function to map between cluster labels and real labels, we are able to produce classification results close to a ResNet18 \cite{resnet} supervised baseline and outperform a feature extraction [last feature layer from ResNet18 combined with KMeans clustering] baseline.

%
%

\vspace{2mm}
\section{Material and Methods}
\vspace{2mm}

\begin{figure}
    \centering
\includegraphics[width=\linewidth]{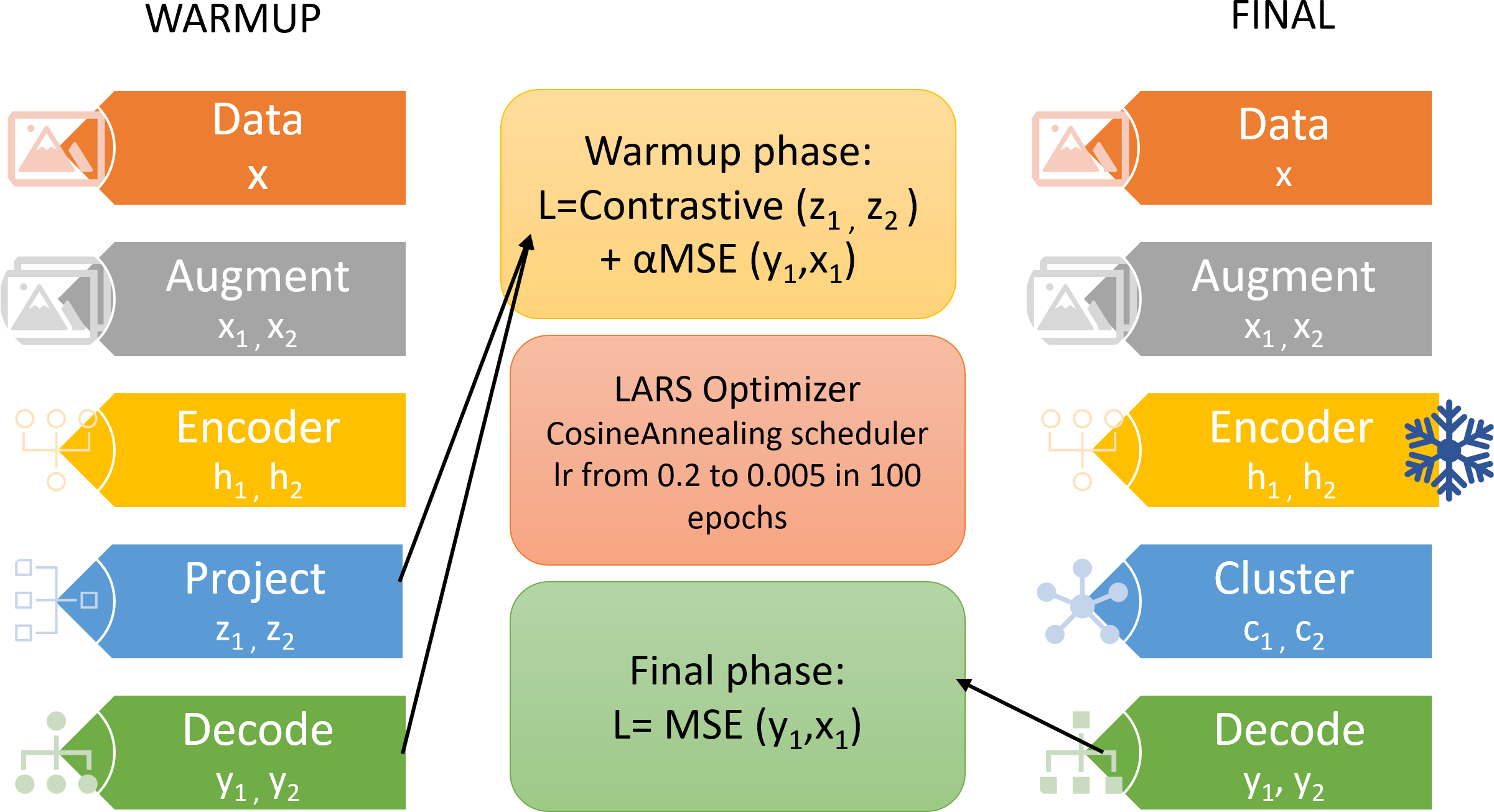}
\caption{ \textit{Architecture overview}}\label{wrap-fig:1}

\end{figure}

Our architecture has the two-fold objective of reconstructing  images while generating a  latent space whose structure separation is driven by contrastive loss. The encoder $e$ (a convolutional network with 3 2D layers,  kernel size=4, stride=2, followed by GELU activations and an average pooling layer (kernel size= 2) and a linear layer that maps the features into a latent space of dimension $128$) maps images $x$ into the latent space $h=e(x)$, while the projector $p$ ( a multilayer perceptron with 3 layers) projects the images into another space $(z=p(h))$  where the similarity function $sim(\cdot,\cdot)$ is computed  as in \cite{simclr}). The encoded representations are then passed to a decoder that learns how to reconstruct the images driven by the reconstruction loss. Parameter values were chosen to halve the image dimensions three times while increasing the receptive field, in order to create filters that process at the entire image feature map before the linear layers.
During the \textit{warmup} (first) phase of training the contrastive loss term

\begin{equation}
    \mathcal{L}_{contrastive}==-\log(e^(\frac{sim(z_i,z_j)}{\tau}/{\sum_{negatives} e^(\frac{sim(z_i,z_k)}{\tau}})
\end{equation}

and a standard mean squared error loss 
 \begin{equation}
 \mathcal{L}=\mathcal{L}_{contrastive}+ \alpha\mathcal{L}_{recon}
\end{equation} 
are combined into the total loss, where $\alpha=0.1$ is a fixed scalar term that encourages learning those features which are potentially also useful for reconstruction. 
All  network weights are updated with the LARS optimizer \cite{lars}. Then, the encoder is frozen and a search for optimal number of clusters is run using KMeans combined with the elbow method. \cite{scikit-learn,yellowbrick}. 
Cluster centroids are then stored in a matrix whose columns represent cluster \textit{prototypes}. 
In the second phase, the decoder is turned into a conditional decoder $d(\cdot)$ by adding layers that process information  from a soft label assignment computed in the latent space. The images are then encoded and their representation are compared to each of the prototypes using cosine distance as similarity metric. The temperature-scaled softmax of the vector of similarities produces a soft label assignment passed to the decoder that will learn how to decode images including this information. The architecture of $d(\cdot)$ is symmetrical with respect to  $e(\cdot)$.

\begin{figure*}[htbp]
    \centering
\includegraphics[width=0.9\linewidth]{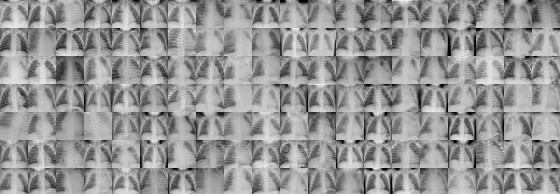}
\caption{ \textit{Examples of data instances from the PneumoniaMNIST dataset}}\label{fig:pneumonia}

\end{figure*}

After obtaining proof-of-concept results on the  MNIST \cite{mnist}  digit images, we validate  our approach on the \textit{pneumoniaMNIST}dataset (2D chest X-Ray images labelled as healthy or affected by pneumonia), part of MedMNIST (\url{https://medmnist.com/}) \cite{medmnist}. We generate \textit{positive} pairs through augmentation, i.e. we randomly apply (with a probability $p=0.5$), rotations of up to 30 degrees, Gaussian blur, Gaussian noise, horizontal flips, and randomly rescaled crops. For all images in the training set we compute the encoded latent representations and their cluster labels. We then map each cluster into the mode of the real labels of items belonging to that specific cluster, to generate a map between the cluster labels and the real labels. The latter map is used to predict the labels for the validation and the test set and compute the performance in a way that is comparable to what is done in a supervised model. An outline of the algorithm is shown below.

\begin{algorithm}[!ht]
  
\begin{algorithmic}
  \FOR{epoch in warmup epochs}{
    \FOR{$x,y$ in dataloader}{

      \STATE$x_1,x_2=$augment($x$)
      \STATE$h_1,h_2=$encode($x_1$,$x_2$)
      \STATE$z_1,z_2=$ project($h_1$,$h_2$) 
      \STATE$y_1, y_2$=decode($h_1$,$h_2$) \\

      \STATE$\mathcal{L}_{sim}$=Contrastive($z_1$,$z_2$) 
      
      \STATE$\mathcal{L}_{recon}$=MSE($x_1$,$y_1$)+MSE($x_2$,$y_2$)
      
      \STATE$\mathcal{L}=\mathcal{L}_{sim}+\alpha\mathcal{L}_{recon}$ \\
      
      \STATE update($e$,$p$,$d$) \\
     \ENDFOR
    \ENDFOR
    
    }
  
  }
  
    \FOR{epoch in(warmup\_epoch,total\_epochs)}{
        \FOR{$x,y$ in dataloader}{

          \STATE$x_1,x_2$=augment($x$)
          \STATE$h_1,h_2$=encode($x_1$,$x_2$)
          \STATE$c_1,c_2$=clusters\_labels($h_1,h_2$)
          \STATE$y_1, y_2$=decode($h_1,h_2,c_1,c_2$) \\

          \STATE$\mathcal{L}$=MSE($x_1,y_1$)+MSE($x_2,y_2$) \\

          \STATE update($d$)
        \ENDFOR
    \ENDFOR
    
    }
  
  }

\end{algorithmic}
\caption{Contrastive autoencoder}

\end{algorithm}

\begin{figure}[h]
    \centering
	\includegraphics[width=1\linewidth]{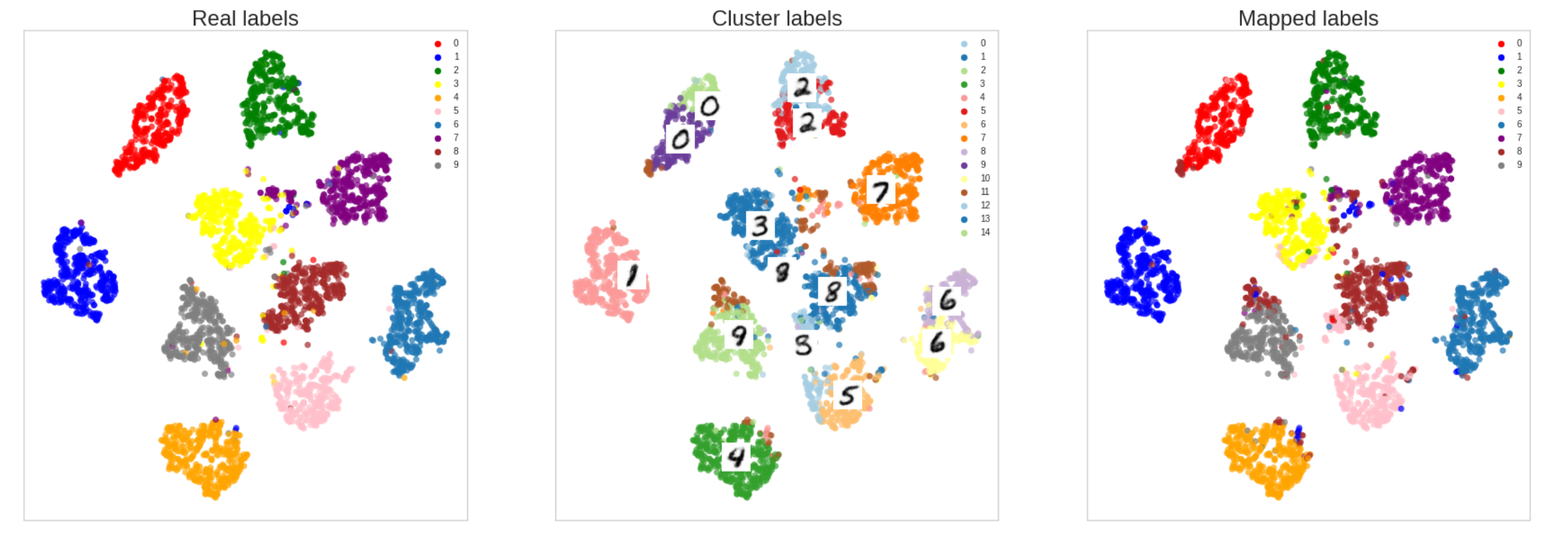}
    \caption{Results on MNIST test dataset.\textbf{A}: real labels, \textbf{B}: cluster labels, \textbf{C}: cluster labels mapped on real labels}
    \label{fig:results}
\end{figure}

\begin{figure}[h]
    \centering
	\includegraphics[width=1\linewidth]{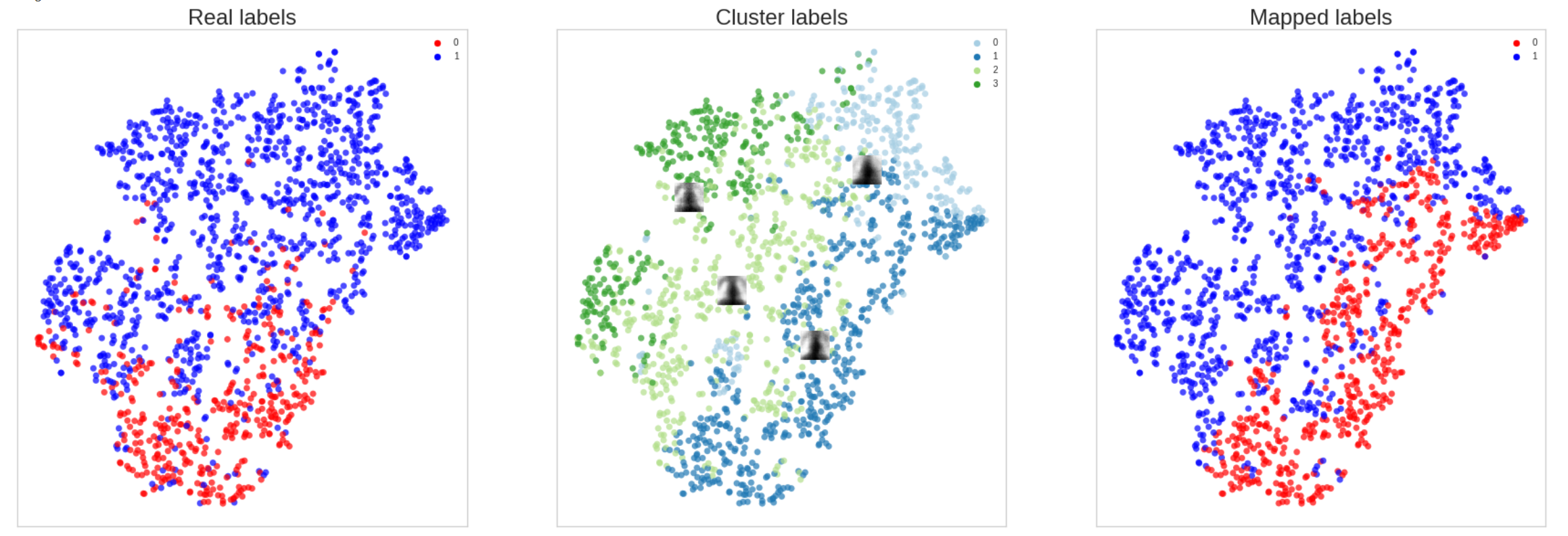}
    \caption{Results on test (PT) and validation (PV) sets of the PneumoniaMNIST dataset. \textbf{A}: real labels, \textbf{B}: cluster labels, \textbf{C}: cluster labels mapped on real labels }
    \label{fig:results}
\end{figure}

We also trained a supervised baseline by performing transfer learning from a ResNet-18 architecture pretrained on ImageNet \cite{imagenet} while changing the number of units in the last fully connected layer to two. This model was trained while freezing all parameters before the avgpool layer (number of epochs: 100, Adam optimizer, learning rate=1e-4). In addition, in order to explore whether embedding the separation in the training results in a performance increases as compared to simply clustering the latent space, we extracted features from the last layer of ResNet18 and performed a KMeans search for optimal number of clusters, followed by a mapping between cluster labels and real labels. The last two experiments served as comparison baselines for the architecture proposed in this paper.
All experiments were ran using Pytorch, 50 warmup epochs and 100 total epochs. The LARS optimizer was used with a StepLR scheduler that linearly increases the learning rate from lr=0.01 to 0.25 (first 10 epochs), after which a cosine annealing scheduler reduces the learning rate to 0.05 in 90 epochs.

\vspace{1mm}
\section{Evaluation}
\vspace{1mm}

For visual evaluation, plot a 2D representation of the clustered latent space using the t-SNE algorithm \cite{tsne}.
Successively, After training the model in an self-supervised manner, the evaluation phase employs part of the available labels are used for performance evaluation in three different approaches, all based on the downstream classification task: a) a statistical mapping approach between cluster labels and real labels, b) the kNN (k-Nearest Neighbours) \cite{knn} algorithm and c) the training of a simple linear layer with categorical cross-entropy as loss function.

In the statistical approach (a), samples drawn from a subportion of the training set are associated with a cluster label which is computed as the index of closest prototype \textit{red}{according to cosine distance.}. Then, the distribution of real labels across each cluster label is computed, and each cluster label is associated with the most frequent real label that occurs in the samples associated with the prototype currently in use. This results in associating each prototype to the mode of the real labels which are most similar to itself, and can be seen as a way to measure how well each prototype capture the key elements that intrinsically define a class. In the inference phase, each sample of the test set is associated to one cluster and a predicted label is generated using this pre-computed map between clusters and real labels.

In the second approach (b), a subset of the training set is used to compute the "memory bank" for the kNN algorithm. In the inference phase, each sample of the test set is compared with every element stored in the memory bank, computing the pairwise distance for every pair. The predicted label is then defined as the mode of the labels of the $k$ closest samples (in our case $k=5$).

In the third approach (c), a linear layer maps from $z$ to the number of possible classes. This linear layer is trained for 200 epochs using the Adam \cite{adam} optimizer with a learning rate of $3*10^{-4}$  over a subportion (20$\%$ of the training set, chosen to mimic a semi-supervised approach in a situation where the number of available labels is low). of the training set, with categorical cross-entropy as loss function.



\vspace{2mm}
\section{Results}
\vspace{2mm}
Results are summarized in Table I.
Since validation datasets were not used for training or model selection, we tested our framework both on the provided validation and test datasets, which have different class proportions and yield different results. Our approach systematically outperforms feature extraction+KMeans in both test and validation set. Importantly, it also performs close to the supervised baseline, even though it is optimised for reconstructing images in an unsupervised manner.
When combined with kNN or a linear classifier our approach generates features which results in even higher performance  as compared to the above performance evaluation approach.
Figure \ref{tab:results} shows a 2D representation of the latent space after encoding for both real and cluster labels. In the latter case, the \textit{prototypes} are also superimposed over the image to demonstrate the reconstruction of mean representatives of each cluster.

\begin{table}[]
    \label{tab:results}

    \begin{tabular}{l@{\hskip 11pt}l@{\hskip 10pt}l@{\hskip 10pt}l@{\hskip 11pt}l@{\hskip 11pt}l@{\hskip 11pt}}

      \bfseries Model & \bfseries Dataset & \bfseries  Accuracy  & \bfseries Precision & \bfseries  Recall\\ \hline
    Resnet18& Pneumonia Test & 0.74 & 0.77 & 0.74  &  \\
      Feature+KM & Pneumonia Test & 0.69 & 0.74 & 0.69 \\
      Our Model (stat) & Pneumonia Test & 0.73 & 0.73 & 0.73 \\ 
      Our Model (kNN) & Pneumonia Test & \textbf{0.84} & 0.85 & 0.83 \\ 
      Our Model (lin) & Pneumonia Test & \textbf{0.84} & 0.80 & 0.88 \\ \hline

Resnet18& Pneumonia Val & \textbf{0.86} & 0.86 & 0.86  \\
      Feature+KM & Pneumonia Val & 0.78 & 0.77 & 0.78 \\
      Our Model & Pneumonia Val & 0.80  & 0.79 & 0.80 \\ 
      Our Model (kNN) & Pneumonia Val & \textbf{0.84} & 0.86 & 0.81 \\ 
      Our Model (lin) & Pneumonia Val & 0.84 & 0.80 & 0.87 \\ \hline
      
      Resnet18 & MNIST & \textbf{0.995} & 0.995 & 0.995 \\
      Feature+KM & MNIST & 0.69 & 0.68 & 0.69  \\
      Our Model & MNIST & 0.90 & 0.91 & 0.90 \\
      Our Model (kNN) & MNIST & \textbf{0.98} & 0.98 & 0.97 \\ 
      Our Model (lin) & MNIST & \textbf{0.98} & 0.97 & 0.98 \\ \hline

      \end{tabular}

        \caption{Results of baseline models and of approach. We evaluated our architecture using three different methods: Statistical (stat), where labels are associated with clusters based on the most frequent label present in a cluster, Nearest Neighbour (kNN) where the class is chosen by majority voting amongst the 5 nearest neighbors in the latent space and a Linear classifier (lin) where we a linear layer was trained on top of the latent features.}

 \end{table}

\vspace{-1mm}
\section{Conclusions}
\vspace{1mm}

This proof of concept study demonstrates the potential of using contrastive learning to encourage latent space separability in autoencoders, and possibly other generative frameworks. 
In benchmark datasets, including medical imaging, our unsupervised stratification method delivers near-equal results (in terms of performance) to supervised baselines, and outperforms the alternative strategy of clustering features extracted from the penultimate layer of the supervised baseline.
The autoencoder framework allows for simultaneous image reconstruction as well as sampling from a specific cluster if used in a variational setting. The method is also apt to refine/redefine existing classes or to completely re-stratify a disease continuum. Current limitations include the need of choosing the number of clusters with KMeans (which could be substituted with e.g. a neural network with learnable weights and a clustering loss) and the need for large batch sizes, which could be foregone through e.g. sampling approaches.

%
%
%
%
\bibliographystyle{splncs04}
\bibliography{samplepaper}
\end{document}